\documentclass[runningheads]{llncs} 
\usepackage{graphicx}
\usepackage{comment}
\usepackage{amsmath,amssymb} 
\usepackage{color}
\usepackage{microtype}
\usepackage{appendix} 

\usepackage[pagebackref=true,breaklinks=true,letterpaper=true,colorlinks,bookmarks=false]{hyperref}

\usepackage{enumitem}
\usepackage{mmstyle}
\usepackage{makecell}
\usepackage{subfig}
\usepackage{dsfont}

\graphicspath{{figures/}}

\begin{document}

\pagestyle{headings}
\mainmatter
\def\ECCVSubNumber{1631}  

\title{Distribution-Balanced Loss for Multi-Label Classification in Long-Tailed Datasets} 

\titlerunning{Distribution-Balanced Loss}
%
\author{Tong Wu\inst{1}~\orcidID{0000-0001-5557-0623} \and
Qingqiu Huang\inst{2}~\orcidID{0000-0002-6467-1634} \and \\
Ziwei Liu\inst{2}~\orcidID{0000-0002-4220-5958} \and
Yu Wang\inst{1}~\orcidID{0000-0001-6108-5157} \and \\
Dahua Lin\inst{2}~\orcidID{0000-0002-8865-7896}}
\authorrunning{T. Wu et al.} 
%
\institute{
    Tsinghua University, Beijing, China\\
    \email{wutong16.thu@gmail.com, yu-wang@mail.tsinghua.edu.cn}\\
\and
    The Chinese University of Hong Kong, Hong Kong, China\\
    \email{zwliu.hust@gmail.com, \{hq016,dhlin\}@ie.cuhk.edu.hk}
}
\maketitle 

\begin{abstract}
    We present a new loss function called Distribution-Balanced Loss for the multi-label recognition problems that exhibit long-tailed class distributions. Compared to conventional single-label classification problem, multi-label recognition problems are often more challenging due to two significant issues, namely the co-occurrence of labels and the dominance of negative labels (when treated as multiple binary classification problems). The Distribution-Balanced Loss tackles these issues through two key modifications to the standard binary cross-entropy loss: 1) a new way to re-balance the weights that takes into account the impact caused by label co-occurrence, and 2) a negative tolerant regularization to mitigate the over-suppression of negative labels. Experiments on both Pascal VOC and COCO show that the models trained with this new loss function achieve significant performance gains over existing methods. Code and models are available at: \url{https://github.com/wutong16/DistributionBalancedLoss}. 

\end{abstract}

\section{Introduction}
\label{sec:introduction}

\begin{figure}[t]
	\centering
	\includegraphics[width=1.0\linewidth]{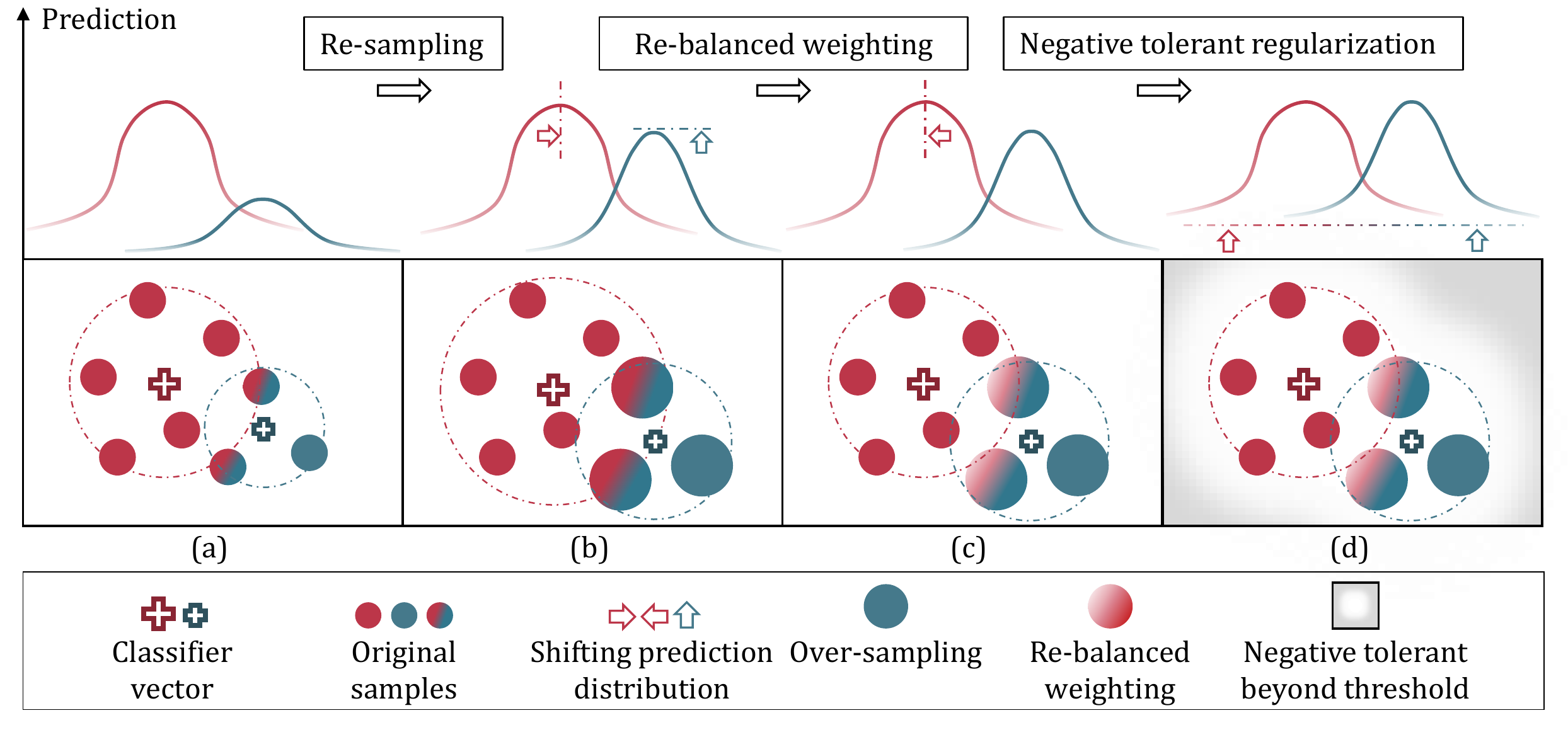}
	\caption{ 
		Our \emph{Distribution-Balanced Loss} performs re-balanced weighting along with re-sampling that takes label co-ocurrence into consideration, and it leverages negative-tolerant regularization to avoid over-suppression of the negative labels caused by the dominance of negative classes in \emph{binary cross entropy (BCE)} 
	}
	\label{fig:overall} 
\end{figure} 

Along with the wide adoption of deep learning, recent years have seen great progress in visual recognition, especially the remarkable breakthroughs in classification tasks. However, mainstream benchmarks are often constructed under two common conditions: 1) all classes have comparable numbers of instances and 2) each instance belongs to a unique class. While providing a clean setting for various studies, this conventional setting conceals a number of complexities that often arise in real-world applications~\cite{liu2019largescale,liu2020open,huang2020caption,huang2018person}. In contrast, the distribution of different object categories typically exhibit a long tail in practical contexts while individual images can generally be associated with multiple semantic labels~\cite{liu2015deep,liu2016deepfashion,huang2020movie,Xiong_2019_ICCV}. Previous works~\cite{japkowicz2002class,horn2017devil,buda2018systematic} have repeatedly shown that such issues can cause substantial performance drop if not appropriately handled.  

A widely adopted approach to multi-label problem is to use binary cross-entropy~\cite{dur2019partial} in the place of the softmax loss, and use class-specific re-weighting to balance the contributions of different classes, e.g. setting the class weights to be inversely proportional to the class sizes. Such simple methods often result in limited improvement, as they fail to take into account the impacts of two important issues, namely \emph{label co-occurrence} and \emph{the dominance of negative labels}.

First, label co-occurrence is very common in natural images. For example, an image that contains unusual concepts, \eg ``tigers'' and ``leopards'', is likely to be also associated with more common labels, \eg ``trees'' and ``river''. Therefore, re-sampling such images may not necessarily result in a more balanced distribution of classes. 
Second, each image is usually associated with a very small fraction of all the classes in the list. Consequently, given an image, most classes are negative. However, the \emph{binary cross entropy (BCE)} loss is designed to be symmetric, where positive and negative classes are treated uniformly. This symmetric formulation in conjunction with the dominant portion of negative classes would lead to over-suppression of the negative side, thus introducing significant bias to the classification boundaries.  
In response to issues above, we propose a new loss function, called \emph{Distribution-Balanced Loss}. This loss function consists of two key modifications to the standard BCE loss: 1) \emph{re-balanced weighting}, which adjust the weights in a way that closes the gap between expected sampling times and actual sampling times, with label co-occurrence taken into account; and 2) \emph{negative-tolerant regularization}, which avoids over-suppression of the negative labels by setting a margin and a re-scaling factor. Experiments on two multi-label recognition benchmarks, \ie~Pascal VOC~\cite{Everingham2015voc} and MS COCO~\cite{lin2014coco}, show that the proposed loss achieves remarkable improvement over previous methods.

\section{Related Work}
\label{sec:related}
Previous works on long-tailed recognition~\cite{wang2017learning,liu2019largescale,kang2020decoupling} mainly follow two directions: re-sampling and cost-sensitive learning. And many efforts have been dedicated to the multi-label classification task. 

\noindent\textbf{Re-sampling.}
To achieve a more balanced distribution, researchers have proposed to either over-sample the minority classes~\cite{shen2016relay,byrd2018effect,buda2018systematic}, or under-sample the majority classes~\cite{japkowicz2002class,he2009learning,buda2018systematic}. The downside of the former is that it might lead to over-fitting on minority classes with duplicated samples, while the latter might weaken feature learning capacity due to omitting a number of valuable instances.
While previous works mainly focus on single label datasets, we extend re-sampling to the multi-label scenario. 

\noindent\textbf{Cost-sensitive Learning.}
Assigning different costs to different training samples is proved to be an effective strategy dealing with imbalanced data.
Typically, researchers apply class-level re-weighting by the proportional inverse of class frequency~\cite{huang2016learning,wang2017learning}, or the square root for smoothing. Recently, Cui~\etal~\cite{cui2019cb} proposed to re-weight by the inverse of effective number of samples, and Cao~\etal~\cite{cao2019ldam} emphasized larger margin for rare classes.
Further, various works adopted sample-level control of cost based on individual properties, \eg example difficulty~\cite{lin2017focal}, estimated Bayesian uncertainty~\cite{khan2019striking}, gradient direction~\cite{ren2018learning}.
Our method applies re-weighting based on class frequency and individual ground truth labels and modifies the loss gradient with a regularization as well for a better optimization.

\noindent\textbf{Multi-label Classification.}
Earlier solutions for multi-label recognition include decomposing it into independent binary classification tasks~\cite{tsoumakas2007overview}, and k-nearest neighbor named ML-kNN~\cite{zhang2007mlknn}, etc.
Recently, many approaches attempted to take label relationships into consideration to better regularize the embedding space. 
CNN-RNN~\cite{wang2016cnnrnn} utilized the RNNs combined with CNN to learn a joint image-label embedding, and Wang~\etal~\cite{wang2017recurrently} took advantages of a spatial transformer layer and long short-term memory(LSTM) units to capture contextual dependencies. There's also a popular trend to model label correlation with graph structure~\cite{lee2018multizero,chen2019mlgcn}.
Our method is based on the widely used binary cross-entropy loss~\cite{dur2019partial} and gains improvement by combining it with re-sampling and re-weighting.

\section{Distribution-Balanced Loss} 
The problem we want to exploit here is how to train a model effectively when training samples follow a long-tailed distribution.
Suppose the dataset we use is $\mathcal{D}= \{(\vx^1, \vy^1), \cdots, (\vx^N, \vy^N)\}$,
where $N$ is the number of training samples and $( \vx^k, \vy^k ),k \in \{1, ... , N\}$ is a sample-label pair.
Let's denote the number of classes as $C$, then we have $\vy^k = [y^k_1, \cdots, y^k_C] \in \{0,1\}^C$.
Let $n_i = \sum_{k=1}^{N} y_i^k$ denote the number of training examples that contain class $i$.
Please note that $N \leq \sum_{i=0}^C n_i$ since a single example can be counted several times for each class it contains.

As we mentioned before, our distribution-balanced loss consists of two components, namely re-balanced weighting and negative-tolerant regularization.
In Sec.~\ref{subsec:weight}, we would introduce the reason why we need a re-balanced weight in long-tailed multi-label classification
and the mathematical derivation of the optimal value of this weight.
In Sec.~\ref{subsec:regularization}, we would demonstrate the over-suppression for negative samples brought by \textit{sigmoid} and how to overcome the problem with our negative-tolerant regularization.
Finally, these two components can be integrated as a unified loss function, \ie distribution-balanced loss, for end-to-end training, which would be shown in Sec.~\ref{subsec:dbloss}.

\subsection{Re-balanced Weighting after Re-sampling} 
\label{subsec:weight}
 
\begin{figure}[t]
    \centering
    \includegraphics[width=\linewidth]{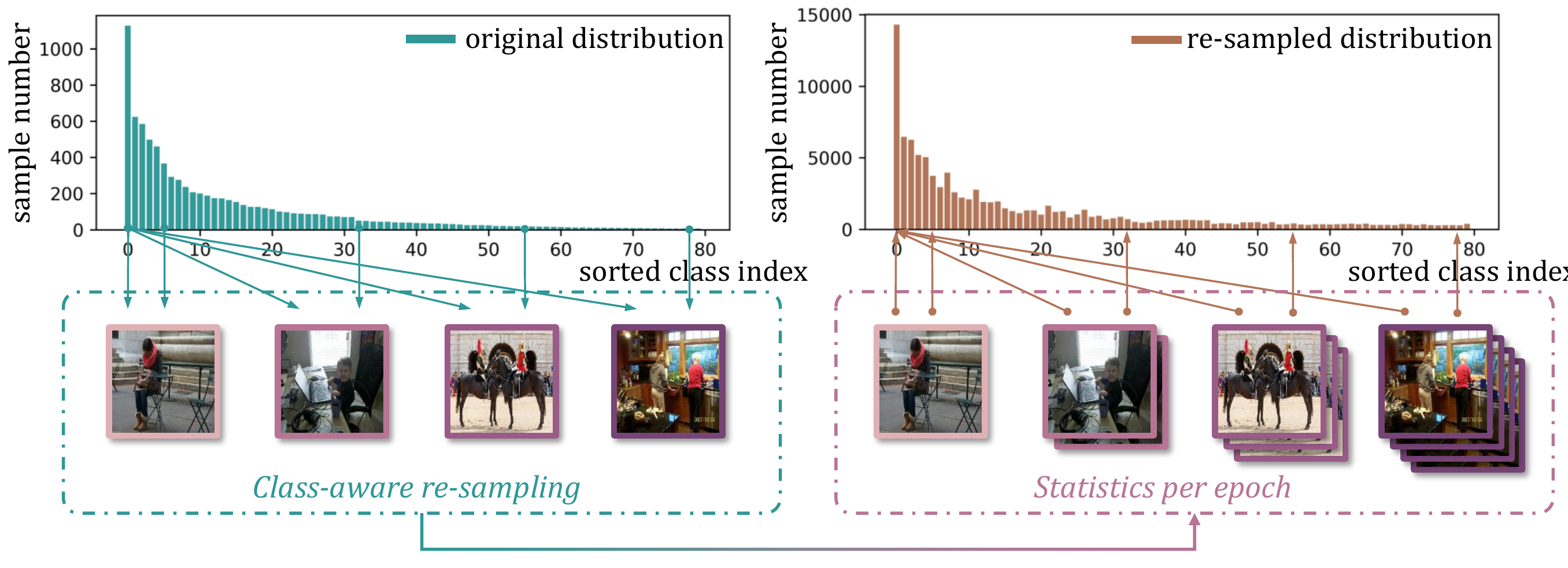} 
    \caption{
        Visualization of the class-aware re-sampling procedure, and the sample number distribution before(left) and after(right) re-sampling. The distribution may not necessarily be balanced due to label co-occurrence, and the unexpected sampling by associated labels introduces inner-class imbalance
    }
    \label{fig:sampling}
\end{figure}

The most common sampling rule is to select each example from training set with equal probability, and the probability of a sampled example containing class $i$ would be $p_i = n_i/N$.
To alleviate the discrepancy of imbalanced sampling probability among classes caused by data distribution, many re-sampling strategies are proposed. One popular strategy is known as class-aware sampling~\cite{shen2016relay,kang2019decoupling,zhou2020BBN}. It first uniformly samples a class from the whole $C$ classes, and then samples an example from the selected class randomly. This process runs iteratively in each training epoch.
Let $N_e$ denote the times for each class to be visited by the iterator in one epoch, which is usually set as $N_e = \max(n_1, \cdots, n_C)$.
In cases of extreme imbalance, $N_e$ can be set smaller to control the data scale in one epoch.

However, in the multi-label scenario, an example usually contains several ground-truth labels, making the selection for classes no longer independent.
That is to say, re-sampling instances from one specific class will inevitably influence the sample numbers of the other classes co-occurring. This leads to the following problems.
First, it induces inner-class imbalance because samples in a class are no longer selected with equal probability.
More importantly, the class imbalance is not necessarily eliminated and may even be exaggerated, the reason for which would be introduced below.

In fact, the numbers of samples for different classes after re-sampling would not follow a uniform distribution as expected.
Here we estimate them using label co-occurence statistics of the original training set. 
Assuming $p(i|j)$ to be the conditional probability of an instance containing label $i$ under the condition of containing label $j$, so that $p(i|j) = n_{i\cap j} / n_j$, where $n_{i\cap j}$ denotes the number of examples that contain both label $i$ and label $j$. 
Therefore, when we randomly choose a class and sample an instance from it, the probability that it contains label $i$ would be shown as Eq.~\ref{eq:pi}.
\begin{equation}
\hat{p}_i = \frac{1}{C}\sum_{j=0}^Cp(i|j) =\frac{1}{C}\sum_{j=0}^C{\frac{n_{i\cap j}}{n_j}}
\label{eq:pi}
\end{equation}

The class distribution after re-sampling is show in Fig.~\ref{fig:sampling}, and the theoratical estimation matches our statistics of data sampled in one epoch during real training procedure. 
According to the distribution, we proposed a re-balanced weighting strategy to overcome the extra imbalance caused by re-sampling.
First, without taking label co-occurrence into consideration, for each instance $k$ and class $i$ with $y^k_i=1$, the expectation of \textbf{C}lass-level sampling frequency can be calculated as $P^C_i(x^k)$ in Eq.~\ref{eq:ei}.
Then given an instance $x^k$ and its corresponding label $y^k$, it is supposed to be repeatedly sampled by each positive class $i$ it contains, thus the expectation of \textbf{I}nstance-level sampling frequency can be estimated as $P^I(x^k)$ in Eq.~\ref{eq:ei}. Correspondingly, we define a re-balancing weight, namely $r^k_i$, to close the gap between expected sampling times and actual sampling times, as shown in Eq.~\ref{eq:rki}.
\begin{equation}
    P^C_i(x^k)=\frac{1}{C}\frac{1}{n_i}, \ P^I(x^k) = \frac{1}{C}\sum_{y^k_i = 1}\frac{1}{n_i}
\label{eq:ei}
\end{equation}
\begin{equation}
    r^k_i = \frac{P^C_i(x^k)}{P^I(x^k)} 
\label{eq:rki}
\end{equation}
\begin{equation}
    \hat{r}^k_i = \alpha + \frac{1}{1 + exp(-\beta \times (r^k_i - \mu))}
    \label{eq:smoothing_func}
\end{equation}

However, the weight elements are sometimes towards zero and may increase the difficulty of optimization.
To make the optimization process stable, we further designed a smoothing function to map $r$ into a proper range of values, which is demonstrated in Eq.~\ref{eq:smoothing_func}.
Here $\alpha$ is an overall lift in weight, while $\beta$ and $\mu$ controls the shape of the mapping function, which rapidly increases near $0$ and goes flat near $1$.

Finally, the loss function, which we name as Re-balanced-BCE, becomes Eq.~\ref{eq:lrbce}, where $z^k$ denotes the output of the classifier.
\begin{equation}
\mathcal{L}_{R-BCE}(x^k, y^k) = \frac{1}{C}\sum_{i=0}^C{ \left[ y^k_i log(1 + e^{-z^k_i}) + (1 - y^k_i)log(1 + e^{z^k_i})\right] \times \hat{r}^k_i}
\label{eq:lrbce}
\end{equation}

What's worth noting is that $\hat{r}^k_i$ is applicable to both positive and negative labels although it was originally deduced from the sampling procedure regarding only the positive ones, 
in order to keep a consistency at class-level.

\subsection{Negative-Tolerant Regularization}
\label{subsec:regularization}

\begin{figure}[t]
    \centering
    \includegraphics[width=\linewidth]{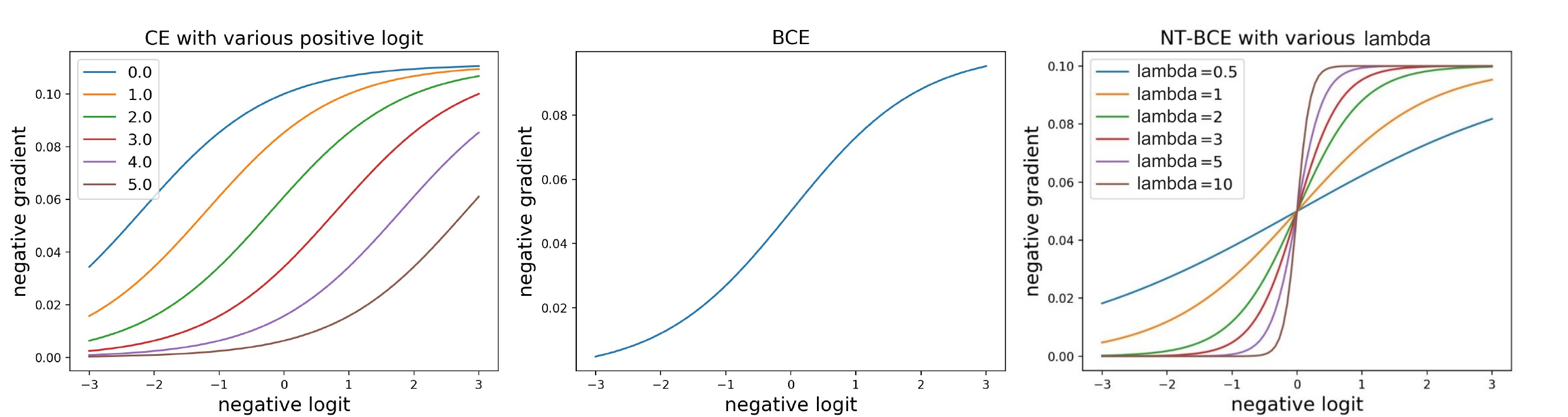}
    \caption{
        Visualization of gradient to a negative logit. (a) The gradient for CE loss can be relatively small with a high positive logit; (b) for BCE loss it's only effected by the negative logit itself which results in continuous suppression; (c) NTR encourages a sharp decrease when the logit is lower than a threshold and slowers down the optimization
    }
    \label{fig:gradient} 
\end{figure}

As mentioned above, \textit{binary cross entropy (BCE)loss}, which is widely used for multi-label classification,
sometimes suffers from over-suppression for negative labels because of the dominance of negative classes. 
To be more specific, BCE considers the recognition task as a series of binary classification tasks, calculating independent class-wise probability with \textit{sigmoid} function.
In contrast, \textit{cross entropy (CE) loss}, which is popular in single-label classification, utilizes \textit{softmax} to emphasize mutual exclusion.
Unlike \textit{softmax} where the optimization step would be rather small once the logit for positive class is much higher than those of negative classes, \textit{sigmoid} treats them independently and encourages the logits of both positive and negative classes to be away from zero in the same gradient declining manner. 
The difference between them can be observed by their gradients shown in Eq.~\ref{eq:gradient} and visualized in Fig.~\ref{fig:gradient}(a)(b).

\begin{equation}
    \left\{
    \begin{aligned}
    \frac{\partial{\mathcal{L}_{CE}(z_j, y)}}{\partial{(z_j)}} = \frac{e^{z_j}}{\sum_{i=0}^Ce^{z_i}}, y_j=0 \\
    \frac{\partial{\mathcal{L}_{BCE}(z_j, y)}}{\partial{(z_j)}} = \frac{1}{C}\frac{e^{z_j}}{1+e^{z_j}}, y_j=0
    \end{aligned}
    \right.
    \label{eq:gradient}
\end{equation}

A straightforward consequence is that the classifiers for the tail classes would over-fit to a limited number of positive samples in the feature space, and meanwhile, they would push a huge number of negative samples away to produce lower logits.
It can be taken as a class-specific over-fitting for the tail categories, which leads to a bad generalization of the classifiers.
As shown in Fig~\ref{fig:overall}(c),
the output distribution becomes sharp and the predictions of the testing samples are easy to be influenced by head classes.

To address the problem, we need a regularization to overcome the over-suppression.
Specifically, the loss by negative logits actually needs a sharp drop once it's optimized to be lower than a threshold so that they won't be continuously suppressed due to a relatively small gradient.
Based on the idea, we propose a negative-tolerant regularization (NTR) by first using a non-zero bias initializaiton to act as the thresholds, and then applying a linear scaling to the negative logits before their calculation in the standard BCE, together with a regularization parameter to constrain the gradient between $0$ and $1$.
The Negative-Tolerant-BCE thus becomes Eq.~\ref{eq:nt-bce}.
\begin{equation}
\mathcal{L}_{NT-BCE}(x^k, y^k) = \frac{1}{C}\sum_{i=0}^C{y^k_i log(1+e^{-(z^k_i-\nu_i)}) + \frac{1}{\lambda}(1 - y^k_i)log(1+e^{\lambda (z^k_i-\nu_i)})}
\label{eq:nt-bce}
\end{equation}

$\lambda$ is the sacle factor that effects the loss gradient as shown in Fig.~\ref{fig:gradient}(c), controling how "tolerant" we are to $z_i$, and $\nu_i$ is a class-specific bias.
The design for $\nu$ is supposed to take intrinsic model bias into consideration.
Concretely, a network trained with imbalanced data is likely to give passive predictions on those tail classes on average, the thresholds for them should correspondingly be lower, assuring that they won't be too easily achieved. 
It shares a similar idea with~\cite{cao2019ldam} that a larger margin is needed for rare classes.
Assuming that we use a fully-connect layer as the classifier, the intrinsic bias of the model can be estimated by minimizing the loss function at the very beginning of training, where the classifiers are randomly initialized, and the dot-product distance between classifier vectors and instance features are at an average of zero.
For a regular BCE loss, considering the bias $b_i$ as the only variable, and assuming the class prior to be $p_i = n_i/N$, we can deduce an approximation of averaged loss by class $i$: 
\begin{equation}
L_{i} = p_i \log(1 + e^{-b_i}) + (1 - p_i) \log(1 + e^{b_i})
\label{eq:li_init}
\end{equation}
\begin{equation}
    \hat{b}_i = -\log(\frac{1}{p_i} - 1), \ \nu_i = -\kappa~\hat{b}_i
\label{eq:value-nu}
\end{equation} 

We minimize Eq.~\ref{eq:li_init} at $\hat{b}_i$, as shown in Eq.~\ref{eq:value-nu}, and use $\kappa$ as a scale factor to get $\nu_i$, which is further applied to Eq.~\ref{eq:nt-bce}. 

\subsection{Distribution-Balanced Loss}
\label{subsec:dbloss}

\begin{figure}[t]
    \centering
    \includegraphics[width=\linewidth]{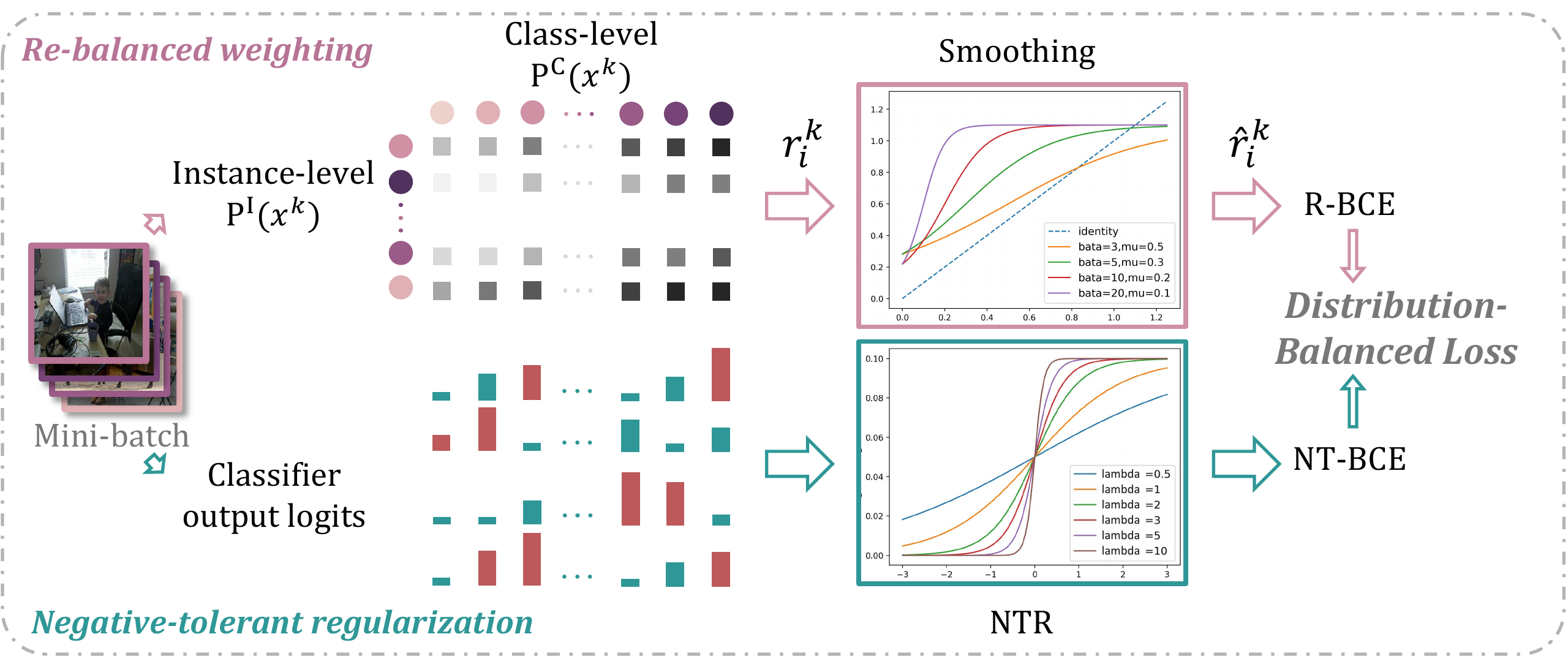}
    \caption{
        Pipeline of the training procedure. Given a mini-batch of instances out of class-aware sampling, 
        the calculation of re-balanced weight is shown in the upper stream, while NTR is shown in the lower. The two techniques are combined to our final distribution-balanced loss
    }
    \label{fig:pipeline}
\end{figure}

So far, R-BCE performs a re-balanced weighting strategy and the weight vector is fixed given an instance, while NT-BCE conducts regularization to the classifier outputs and affects the training by modifying the loss gradient. They can be naturally integrated as a unified loss function for end-to-end training, as shown in Fig.~\ref{fig:pipeline}, and we finally get our distribution-balanced loss as Eq.~\ref{eq:db-loss}.
\begin{equation}
    \mathcal{L}_{DB}(x^k, y^k) = \frac{1}{C}\sum_{i=0}^C \hat{r}^k_i \left[ y^k_i log(1+e^{-(z^k_i-\nu_i)}) + \frac{1}{\lambda}(1 - y^k_i)log(1+e^{\lambda (z^k_i-\nu_i)}) \right]
    \label{eq:db-loss}
\end{equation}

DB loss helps to smooth the distribution of the classifier outputs especially for those tail classes. It achieves superior performance in muti-label datasets with a long-tailed distribution, as will be validated in Sec.~\ref{sec:experiment}.

\section{Experiments}
\label{sec:experiment}

\subsection{Datasets}
The proposed \textit{Distribution Balanced Loss} is analyzed on artificially created long-tailed versions of two multi-label image recognition benchmarks, named VOC-MLT and COCO-MLT, respectively. 
They're subsets sampled from the original datasets by \textit{pareto distribution} $pdf(x) = \alpha \frac{x_{min}^\alpha}{x^{\alpha + 1}}$ following~\cite{liu2019largescale}. $\alpha$ controls how fast data scale decays.
Regarding the interaction of sampling among classes, we construct the datasets in a head-to-tail manner so that we can strictly constrain the scale of tail classes:
we first rank all the classes by $\hat{p}_i$ calculated with original data, and for each class $i$ from head to tail, we \textit{add} or \textit{eliminate} instances that contain class $i$ from the subset by referring to the expected distribution.
Details on the construction of VOC-MLT and COCO-MLT can be found in the appendix.

\noindent\textbf{VOC-MLT.} 
We construct the long-tailed version of VOC~\cite{Everingham2015voc} from its 2012 train-val set, with the power parameter $\alpha=6$.
It contains 1,142 images from 20 classes, with a maximum of 775 images per class and a minimum of 4 images per class.
We evaluate the performance on VOC2007 test set with 4952 images.
 
\noindent\textbf{COCO-MLT.}
The long-tailed version of MS COCO-2017~\cite{lin2014coco} is created with the same $\alpha$, containing 1,909 images from 80 classes. 
The maximum of training number per class is 1,128 and the minimum is 6. 
We use the test set of COCO2017 with 5,000 for evaluation. 
What's worth noting is that the test set of COCO and VOC are not perfectly balanced, they share a similar distribution with the original train set. But the ranking of sample scale per-class of both the original train and test set is roughly consistent with the long-tailed version.

\subsection{Experimental Settings}

\noindent\textbf{Evaluation Metrics.}
Following~\cite{liu2019largescale}, we split the classes into three groups by the number of their training examples: \textit{head} classes each contains over 100 samples, \textit{medium} classes each has between 20 and 100 samples, and \textit{tail} classes with under 20 samples each. 
We evaluate mean average precision(mAP) for all the classes, and we also report mAP for each subset to observe how the techniques effect on them.

\noindent\textbf{Comparing Methods.}
We compare our methods with several state-of-the-art techniques dealing with multi-label classification or long-tailed recognition. We also report the results of their effective combinations for fair comparison.
The standard binary cross entropy loss with sigmoid function is used or modified by all the methods. 
The compared methods include:
(1) Empirical risk minimization: The plain model with all the examples having the same weight and sampling probability. 
(2) Re-weighting(RW): we perform a smooth version of re-weighting to be inversely proportional to the square root of class frequency, and we normalize the weights to be between 0 and 1 in a mini-batch.
(2) Re-sampling(RS)~\cite{shen2016relay}: we use class-aware re-sampling without extra tricks as a baseline, and we also evaluate the combination of RS and other techniques in comparison.
(3) ML-GCN~\cite{chen2019mlgcn}: a recently proposed method by for multi-label classification with \textit{graph convolutional network}(GCN).
(4) OLTR~\cite{liu2019largescale}: set a new benchmark for single-label long-tailed recognition, so we make modifications to their method and adapt it into multi-label case. 
(5) Focal loss~\cite{lin2017focal}: we use $\gamma=2$ with a balance parameter of 2 for focal loss.
(6) Class-balanced loss(CB)~\cite{cui2019cb}: a class-wise re-weighting guided by the effective number of each class $E_n=(1-\beta^n)/(1-\beta)$. 
(7) Label-distribution-aware margin loss(LDAM)~\cite{cao2019ldam}: a recently proposed margin-loss which is proved to be effective for softmax classifier.

\noindent\textbf{Implementation Details.}
We adopt Resnet50~\cite{he2016deep} pretrianed on ImageNet~\cite{deng2009imagenet} as backbone feature extractor, followed by global average pooling and a $2048 \times 256$ fully connection(FC) layer to obtain image-level features. The final classifier is a $256 \times C$ FC layer which outputs the logits. 
The input images are organized with a batch size of 32, randomly cropped and resized to $224\times224$ together with standard data augmentation.
We use SGD with momentum of 0.9 and weight decay of $1\times 10^{-4}$ as our optimizer, and we also use linear warm-up learning rate schedule~\cite{goyal2017warmup} for the first 500 iterations with a ratio of $\frac{1}{3}$.
Training not combined with re-sampling is trained for 80 epochs with an initial learning rate of 0.02, which decays by a factor of 10 after 55 and 70 epochs, respectively.
Re-sampling enhanced methods are trained for 8 epochs with the same learning rate initialization decaying step, and it decays after 5 and 7 epochs, respectively.
We use the class-aware re-sampling~\cite{shen2016relay} and the times that the iterator visits each class in one epoch is set as $N_{max}/4$ where $N_{max}$ denotes the maximum number of training samples per class. 
Once $N_e<N_{max}$, it actually controls how we over-sample the tail classes and under-sample the head classes with no tail ones co-occurring. 
The experiments are implemented in PyTorch.

 \begin{table}[t]
 	\centering
 	\caption{
 		Experimental results of mAP by our methods and other comparing approaches on VOC-MLT and COCO-MLT.
 		We evaluate the results on the whole class set and the three subsets, respectively
 	}
 	\resizebox{\linewidth}{!}{
 		\begin{tabular}{c|c|c|c|c||c|c|c|c} 
 			\hline
 			\multicolumn{1}{c|}{\textbf{Datasets}}& \multicolumn{4}{c||}{VOC-MLT} & \multicolumn{4}{c}{COCO-MLT} 
 			\\ \hline
 			\multicolumn{1}{c|}{\textbf{Methodss}}& total & head  & medium & tail  & total & head  & medium & tail  \\ \hline
 			ERM         & 70.86 & 68.91 & 80.20  & 65.31 & 41.27 & 48.48 & 49.06  & 24.25 \\
 			RW & 74.70 & 67.58 & 82.81  & 73.96 & 42.27 & 48.62 & 45.80  & 32.02 \\
 			Focal Loss~\cite{lin2017focal}  & 73.88 & 69.41 & 81.43  & 71.56 & 49.46 & 49.80 & 54.77  & 42.14 \\
 			RS~\cite{shen2016relay} & 75.38 & 70.95 & 82.94 & 73.05 & 46.97 & 47.58 & 50.55 & 41.7 \\
 			RS-Focal    & 76.45 & 72.05 & 83.42  & 74.52 & 51.14 & 48.90 & 54.79  & 48.30 \\
 			ML-GCN~\cite{chen2019mlgcn}  & 68.92 & 70.14 & 76.41 & 62.39 & 44.24 & 44.04 & 48.36  & 38.96 \\
 			OLTR~\cite{liu2019largescale}  & 71.02 & 70.31 & 79.80 & 64.95 & 45.83 & 47.45 & 50.63  & 38.05 \\
 			LDAM~\cite{cao2019ldam}    & 70.73 & 68.73 & 80.38  & 69.09 & 40.53 & 48.77 & 48.38  & 22.92 \\
 			CB-Focal~\cite{cui2019cb}    & 75.24 & 70.30 & 83.53  & 72.74 & 49.06 & 47.91 & 53.01  & 44.85 \\\hline
 			R-BCE          & 76.34 & 71.40 & 82.76  & 75.22 & 49.43 & 48.77 & 53.00  & 45.33 \\
			R-BCE-Focal    & 77.39 & 72.44 & 83.16  & 76.77 & 52.75 & 50.20 & 56.52  & 50.02 \\
			DB             & 78.65 & 73.16	& 84.11	& 78.66	& 52.53	& 50.25	& 56.33	& 49.54  \\
 			DB-Focal  & ~\textbf{78.94}~ & ~\textbf{73.22}~ & ~\textbf{84.18}~  & ~\textbf{79.30}~ & ~\textbf{53.55}~ & ~\textbf{51.13}~ & ~\textbf{57.05}~ & ~\textbf{51.06} \\
 			\hline
 		\end{tabular}
 	}
 \label{table:main_map}
 \end{table}

\subsection{Benchmarking Results} 

\noindent\textbf{VOC-MLT.}
VOC-MLT contains 6, 6, and 8 classes for the head, medium, and tail classes, respectively. 
We adjusted the hyper-parameters in other methods so that they currently work best in our dataset.
For our method, we choose $\alpha=0.1$, $\beta=10$, and $\mu=0.3$ for the smoothing function during re-balanced weighting.
And we set $\lambda=5, \kappa=0.05$ for NTR.
The experimental results compared with other traditional and state-of-the-art approaches can be seen in Table~\ref{table:main_map}.
What's worth noting is that unlike COCO-MLT whose tail classes always co-occur with head classes, the tail classes for VOC usually appear as single-label, which lower the complexity and difficulty for the classification task on them, bringing higher performance for the tail that even outperforms the head.
This character also alleviate the defect of inner-class imbalance cause by head-tail connection and notice that after regular re-sampling, all subsets including the head have an improvement, especially the tail classes.
Comparing with the best baseline, re-sampling trained with focal loss (RS-Focal), our re-balanced weighting strategy gains further improvement by about $1.0\%$ in total mAP, with $0.4\%$ and $2.2\%$ for head and tail classes and drop $0.3\%$ for medium classes.
Our final DB-Loss further achieves remarkable improvements by $1.6\%$ compared with R-BCE, and by $0.8\%$, $1.0\%$ and $2.5\%$ for the three subsets, respectively. It can be seen that NTR is especially beneficial for the tail classes.

\begin{figure}[t]
    \centering
    \includegraphics[width=\linewidth]{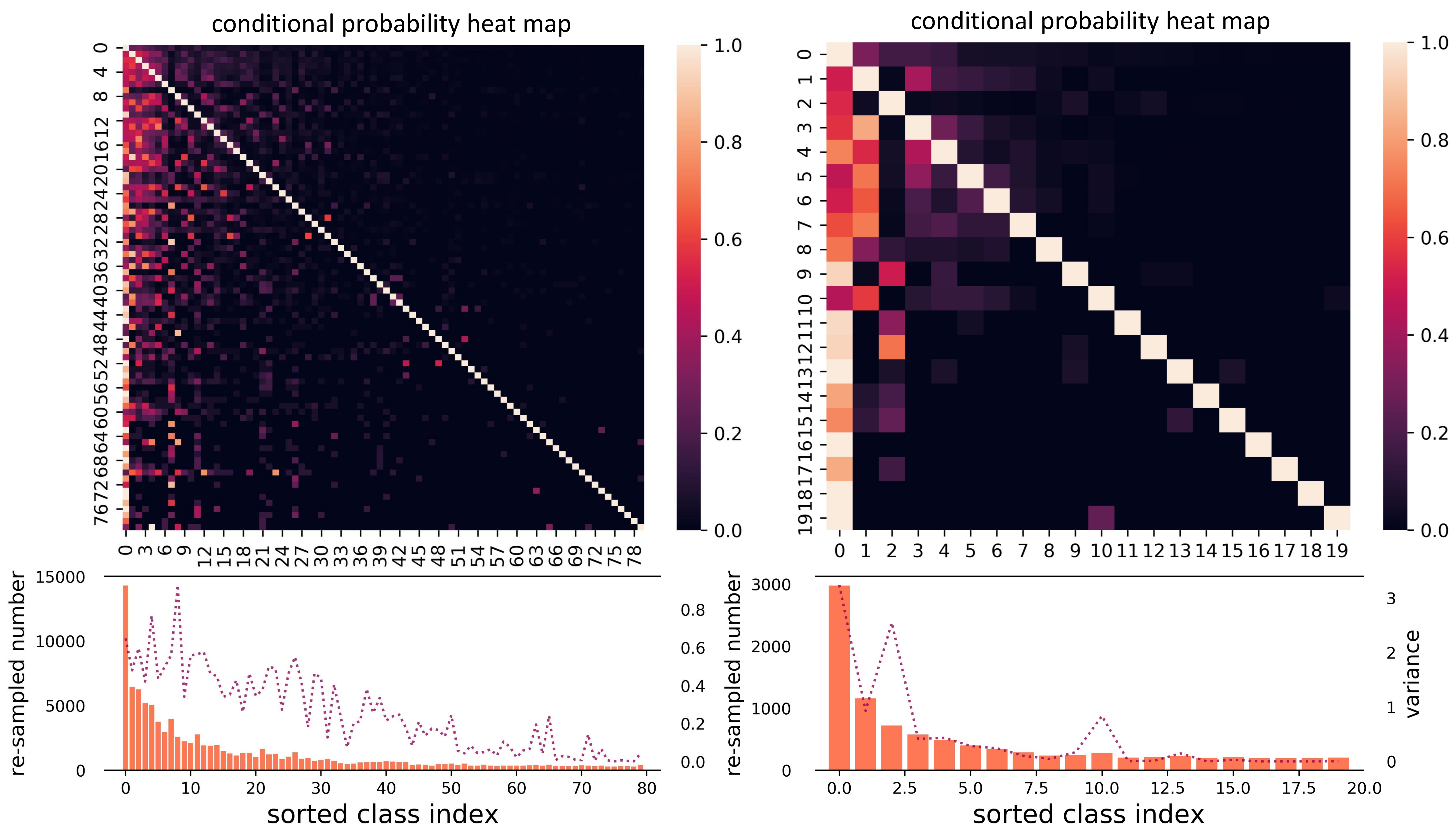}
    \caption{
        In the matrix heatmap, the element in the $i^{th}$ column, $j^{th}$ row represents the conditional probability of class existence, $p(i|j)$, which is usually higher for head and medium classes.
        The histograms below show the data distribution after re-sampling, confirming that the imbalance is not eliminated by re-sampling.
        The sampling frequency for each instance is different, and the line chart shows the variance within each class. The high variance at the head indicates inner-class imbalance
        }
    \label{fig:heatmap_and_distribution}
\end{figure}

\noindent\textbf{COCO-MLT.}
The whole 80 classes of COCO-MLT are split into 22, 33, and 25 classes for the head, medium, and tail classes, respectively.
We choose $\alpha=0.1$, $\beta=10$, and $\mu=0.2$ for the smoothing function during re-balanced weighting.
And we set $\lambda=2, \kappa=0.05$ for NTR.
The experimental results compared with other traditional and state-of-the-art approaches can be seen in Table~\ref{table:main_map}.
COCO-MLT has a heavy head-tail connection, \ie some tail classes has a $100\%$ probability of oc-curring with certain head classes, as shown in Fig~\ref{fig:heatmap_and_distribution} A direct result of class-aware re-sampling is a sharp rise of mAP to the tail classes, while the performance for head classes drops by about $0.9\%$. With re-balanced weighting, the negative effect on the head classes is fixed and mAP for head, medium, and tail classes all have an improvement.
With focal loss combined with either BCE trained re-sampling or R-BCE trained re-sampling, we see an extra improvement.
Using DB-Loss would further bring an average improvement of about $0.8\%$, and brings up mAP for tail classes by about $1.0\%$.

\subsection{Ablation Study}

\noindent\textbf{Visualization of the Imbalance caused by Re-sampling.}
We visualize the conditional probability matrix that reveals label co-occurrence relationship in Fig~\ref{fig:heatmap_and_distribution}. As can be seen that the most frequently appearing classes usually have the highest co-existing probability on the conditionon of other classes. This makes them repeatedly sampled, and the imbalance is not eliminated after re-sampling.
We also roughly estimate the inner-class imbalance by the variance of normalized sampling times: 
for each class and each of the instances containing it, we calculate its expected sampling times. We normalize them within a class to a mean value of 1, and the variance of the normalized sampling times is calculated, as shown in Fig~\ref{fig:heatmap_and_distribution}. 
Variance can roughly represent the extent of imbalance in sampling.
Classes with high variance gain little or negative increment in mAP despite heavy sampling on them. 
A more precise and complicated cooperation of sampling variance and data scale is out of the scope of this paper, which can be reserved as future work.
\begin{figure}[t]
    \centering
    \includegraphics[width=\linewidth]{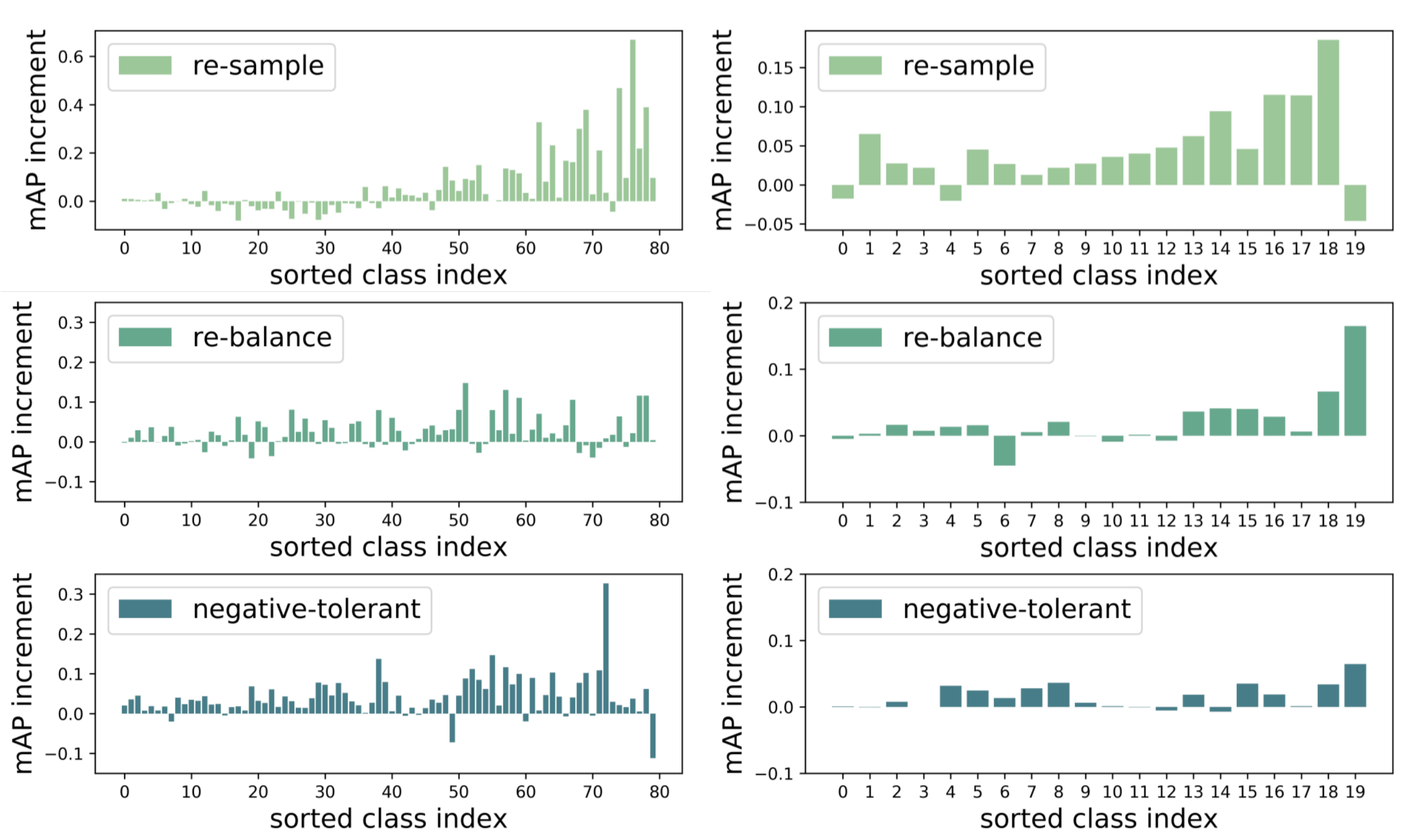}
    \caption{
    Corresponding to the four training strategies \ie ERM, regular class-aware re-sampling, re-sampling with R-BCE and the final DB-Loss, we show the per-class mAP increments between each two steps to evaluate our pipeline piecemeal, presented from left to right.
    Result on COCO-MLT is in the left and VOC-MLT is in the right
    } 
    \label{fig:map_increment}
\end{figure}

\noindent\textbf{Step-wise Evaluation of our Framework.}
We perform a step-wise evaluation on the test set by showing mAP increment per-class to have a better understanding of how re-balanced weighting and negative-tolerant regularization work on different parts of the dataset distribution. 
As shown in Fig.~\ref{fig:map_increment} and mentioned above, regular re-sampling is not friendly to head and medium classes, with little or negative increment. While using re-balanced weighting has a general improvement among classes and amend performance drop by regular re-sampling. 
Negative-tolerant regularization also benefits a wide range of classes, and it leads to a remarkable improvement for tail classes as we expect, indicating an improved generalization ability after the suppression of negative samples is relaxed.

\noindent\textbf{The Combination of Re-sampling and Various Re-weighting Methods.}
Re-sampling and traditional re-weighting methods based on the original distribution share a similar thought of drawing more importance to the rare classes, and they're usually performed at instance level. As a result, the combinations of them are at risk of redundancy: the head classes are over-ignored and the tail classes are over-emphasized. 
Our re-balanced weight is also calculated from the training distribution and it's designed to fine-tune and enhance re-sampling rather than doing repetitive jobs.
So we combine the traditional re-weighting methods with re-sampling for comparison as performed in Table~\ref{table:RS_plus_RW}. Our superior performance shows the benifit of applying R-BCE to re-sampling.

\begin{table}[t]
    \centering
    \caption{
    	Experimental results on re-sampling combined with several re-weighting techniques. CB loss with focal is reported by~\cite{cui2019cb} to perform better, so all the other techniques are enhanced with focal loss for fair comparison
    }
    \resizebox{\linewidth}{!}{
    \begin{tabular}{c|c|c|c|c||c|c|c|c} 
        \hline
    \multicolumn{1}{c|}{\textbf{Datasets}}& \multicolumn{4}{c||}{VOC-MLT} & \multicolumn{4}{c}{COCO-MLT} 
    \\ \hline
    \multicolumn{1}{c|}{\textbf{Methodss}}& total & head  & medium & tail  & total & head  & medium & tail  \\ \hline
    RS~\cite{shen2016relay}          & 75.38 & 70.95 & 82.94  & 73.05 & 47.7  & 46.44 & 51.93  & 43.23 \\\hline
    RS-Focal~\cite{lin2017focal}    & 76.45 & 72.05 & 83.42  & 74.52 & 51.14 & 48.9  & 54.79  & 48.30 \\
    RS+RW-Focal~\cite{mikolov2013distributed,mahajan2018exploring} & 71.96 & 63.14 & 81.09  & 71.73 & 49.07 & 47.80 & 52.09  & 46.20 \\ 
    RS+CB-Focal~\cite{cui2019cb} & 75.24 & 70.30 & \textbf{83.53}  & 72.74 & 50.07 & 48.45 & 54.03  & 48.28 \\\hline
    R-BCE-Focal & ~\textbf{77.39}~ & ~\textbf{72.44}~ & 83.16  & ~\textbf{76.77}~ & ~\textbf{52.75}~ & ~\textbf{50.2}~  & \textbf{56.52}  & ~\textbf{50.02}~ \\\hline 
    \end{tabular}
	}
    \label{table:RS_plus_RW}
   \end{table}

\subsection{Further Analysis}

\noindent\textbf{The Effect of Hyper-parameters of Smoothing Function.}
The smoothing function Eq.~\ref{eq:smoothing_func} has three hyper-parameters, where $\alpha$ applies an overall lift in weight,  $\beta$ and $\mu$ control the shape of the mapping function. We report the results of $\beta$ with $\mu=0.2$ fixed as shown in Fig.~\ref{eq:smoothing_func}. 
\begin{figure}[t]
    \subfloat[\label{fig:smoothing_beta}]{\includegraphics[width=0.49\linewidth]{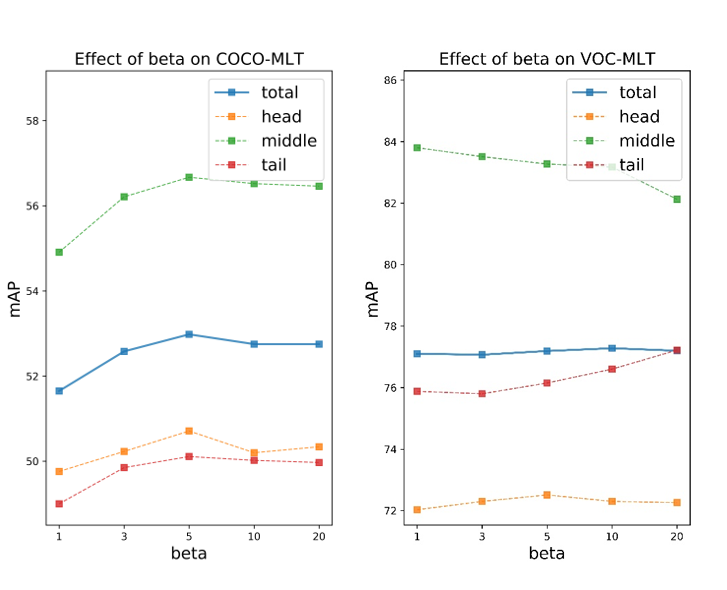}} \hfill
    \subfloat[\label{fig:nt_lambda}]{\includegraphics[width=0.49\linewidth]{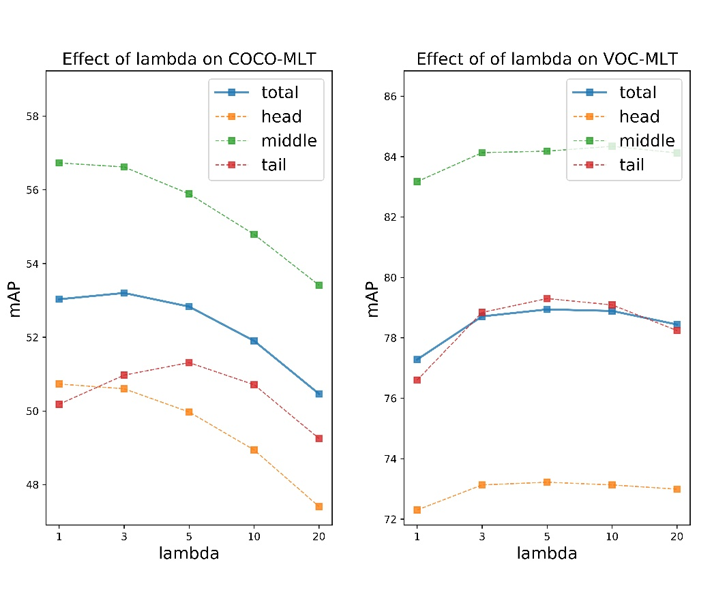}}
    \caption{
        (a). We show how mAP is effected by $\beta$ independently. The total mAP has a peak at around $5<\beta<10$ for both datasets. And we observe an inverse tendency of results for different subsets when $\beta>5$.
        (b). We show how mAP is effected by $\lambda$ independently. We can observe a peak for both datasets away from $\lambda=1$. COCO is more sensitive towards it and we finalize at choosing $\lambda=2$ in the main experiment
    } 
\end{figure}

\noindent\textbf{The Effect of $\lambda$ in Negative-tolerant Regularization.}
To understand how $\lambda$ and $\nu$ of Eq.~\ref{eq:nt-bce} affect the results independently, we first fix $\nu=0$ the same as in the main experiments and change $\lambda$ in a large range, and then we fix $\lambda=2,5$ for COCO-MLT and VOC-MLT, respectively, and change $\nu$.
The effect of $\nu$ is relatively small, and we would report the results in the supplementary material.
Here in Fig.~\ref{fig:nt_lambda}, we observe that it performs the best at $\lambda\in [5,10]$ for VOC, with an improvement of about $2.5\%$ for the tail classes, and $1\%$ for head and medium classes. In COCO, head and medium classes are slightly affected when $\lambda<3$ and tail classes have an improvement of about $1\%$ at around $2<\lambda<3$.
What's worth noting is that, by adding the same form of regularization to positive logits, the results slightly drop as expected.

\noindent\textbf{Group-wise Analysis.}
Medium classes always have a better mAP on average than both the head and tail classes.
The reason for this may be that, medium classes neither suffer from the over-fitting problem as tail classes due to insufficient training samples, nor do they get hurt from the imbalance induced by re-sampling.
Another explanation for this is that the average number of classes an instance has is gradually reduced from the head classes to the tail. 
This indicates a lower complexity and difficulty for the recognition task. 
For instance, quite a number of the training samples for the tail classes of VOC-MLT have only one ground-truth label. Phenomena led by this is that the defect of re-sampling is relieved and the mAP performance of the tail classes surprisingly outperforms head classes by a margin.


\section{Conclusion}
\label{sec:conclusion}

In this work, we propose a simple yet powerful loss function, Distribution-Balanced Loss, to tackle the multi-label long-tailed recognition problem. Multi-label long-tailed recognition problem has two intrinsic challenges, namely the co-occurrence of labels and the dominance of negative labels (when treated as multiple binary classification problems). To tackle these two obstacles, the Distribution-Balanced Loss consists of two key ingredients: 1) a new way to rebalance the weights that takes into account the impact caused by label co-occurrence, and 2) a negative tolerant regularization to mitigate the over-suppression of negative labels. Extensive experiments on both Pascal VOC and COCO validate the effectiveness of the Distribution-Balanced Loss to tackle multi-label long-tailed visual data. The models trained with our new loss function achieve significant performance gains over existing methods, which we believe will serve as a strong baseline for future research.

\noindent\textbf{Acknowledgements.}
This work is partially supported by the SenseTime Collaborative Grant on Large-scale Multi-modality Analysis (CUHK Agreement No. TS1610626 \& No. TS1712093), the General Research Fund (GRF) of Hong Kong (No. 14236516 \& No. 14203518), and Innovation and Technology Support Program (ITSP) Tier 2, ITS/431/18F. Correspondence to Ziwei Liu.

\bibliographystyle{splncs04}
\bibliography{egbib} 

\begin{thebibliography}{10}
\providecommand{\url}[1]{\texttt{#1}}
\providecommand{\urlprefix}{URL }
\providecommand{\doi}[1]{https://doi.org/#1}

\bibitem{buda2018systematic}
Buda, M., Maki, A., Mazurowski, M.A.: A systematic study of the class imbalance
  problem in convolutional neural networks. Neural Networks  \textbf{106},
  249--259 (2018)

\bibitem{byrd2018effect}
Byrd, J., Lipton, Z.C.: What is the effect of importance weighting in deep
  learning? arXiv preprint arXiv:1812.03372  (2018)

\bibitem{cao2019ldam}
Cao, K., Wei, C., Gaidon, A., Arechiga, N., Ma, T.: Learning imbalanced
  datasets with label-distribution-aware margin loss. In: Advances in Neural
  Information Processing Systems (NIPS). pp. 1565--1576 (2019)

\bibitem{chen2019mlgcn}
Chen, Z.M., Wei, X.S., Wang, P., Guo, Y.: Multi-label image recognition with
  graph convolutional networks. In: Proceedings of the IEEE Conference on
  Computer Vision and Pattern Recognition (CVPR). pp. 5177--5186 (2019)

\bibitem{cui2019cb}
Cui, Y., Jia, M., Lin, T.Y., Song, Y., Belongie, S.: Class-balanced loss based
  on effective number of samples. Proceedings of the IEEE Conference on
  Computer Vision and Pattern Recognition (CVPR)  (2019)

\bibitem{deng2009imagenet}
Deng, J., Dong, W., Socher, R., Li, L.J., Li, K., Fei-Fei, L.: Imagenet: A
  large-scale hierarchical image database. In: Proceedings of the IEEE
  Conference on Computer Vision and Pattern Recognition (CVPR). pp. 248--255.
  Ieee (2009)

\bibitem{dur2019partial}
Durand, T., Mehrasa, N., Mori, G.: Learning a deep convnet for multi-label
  classification with partial labels. In: Proceedings of the IEEE Conference on
  Computer Vision and Pattern Recognition (CVPR). pp. 647--657 (2019)

\bibitem{Everingham2015voc}
Everingham, M., Eslami, S.M.A., Van~Gool, L., Williams, C.K.I., Winn, J.,
  Zisserman, A.: The pascal visual object classes challenge: A retrospective.
  International Journal of Computer Vision (IJCV)  \textbf{111}(1),  98--136
  (2015)

\bibitem{goyal2017warmup}
Goyal, P., Doll{\'a}r, P., Girshick, R., Noordhuis, P., Wesolowski, L., Kyrola,
  A., Tulloch, A., Jia, Y., He, K.: Accurate, large minibatch sgd: Training
  imagenet in 1 hour. arXiv preprint arXiv:1706.02677  (2017)

\bibitem{he2009learning}
He, H., Garcia, E.A.: Learning from imbalanced data. IEEE Transactions on
  Knowledge and Data Engineering  \textbf{21}(9),  1263--1284 (2009)

\bibitem{he2016deep}
He, K., Zhang, X., Ren, S., Sun, J.: Deep residual learning for image
  recognition. In: Proceedings of the IEEE Conference on Computer Vision and
  Pattern Recognition (CVPR). pp. 770--778 (2016)

\bibitem{horn2017devil}
Horn, G.V., Perona, P.: The devil is in the tails: Fine-grained classification
  in the wild. arXiv preprint arXiv:1709.01450  (2017)

\bibitem{huang2016learning}
Huang, C., Li, Y., Change~Loy, C., Tang, X.: Learning deep representation for
  imbalanced classification. In: Proceedings of the IEEE Conference on Computer
  Vision and Pattern Recognition (CVPR). pp. 5375--5384 (2016)

\bibitem{huang2018person}
Huang, Q., Liu, W., Lin, D.: Person search in videos with one portrait through
  visual and temporal links. In: Proceedings of the European Conference on
  Computer Vision (ECCV). pp. 425--441 (2018)

\bibitem{huang2020movie}
Huang, Q., Xiong, Y., Rao, A., Wang, J., Lin, D.: Movienet: A holistic dataset
  for movie understanding. In: Proceedings of the European Conference on
  Computer Vision (ECCV) (2020)

\bibitem{huang2020caption}
Huang, Q., Yang, L., Huang, H., Wu, T., Lin, D.: Caption-supervised face
  recognition: Training a state-of-the-art face model without manual
  annotation. In: Proceedings of the European Conference on Computer Vision
  (ECCV) (2020)

\bibitem{japkowicz2002class}
Japkowicz, N., Stephen, S.: The class imbalance problem: A systematic study.
  Intelligent Data Analysis  \textbf{6}(5),  429--449 (2002)

\bibitem{kang2020decoupling}
Kang, B., Xie, S., Rohrbach, M., Yan, Z., Gordo, A., Feng, J., Kalantidis, Y.:
  Decoupling representation and classifier for long-tailed recognition. In:
  International Conference on Learning Representations (ICLR) (2020)

\bibitem{kang2019decoupling}
Kang, B., Xie, S., Rohrbach, M., Yan, Z., Gordo, A., Feng, J., Kalantidis, Y.:
  Decoupling representation and classifier for long-tailed recognition. In:
  Eighth International Conference on Learning Representations (ICLR) (2020)

\bibitem{khan2019striking}
Khan, S., Hayat, M., Zamir, S.W., Shen, J., Shao, L.: Striking the right
  balance with uncertainty. Proceedings of the IEEE Conference on Computer
  Vision and Pattern Recognition (CVPR)  (2019)

\bibitem{lee2018multizero}
Lee, C.W., Fang, W., Yeh, C.K., Frank~Wang, Y.C.: Multi-label zero-shot
  learning with structured knowledge graphs. In: Proceedings of the IEEE
  Conference on Computer Vision and Pattern Recognition (CVPR). pp. 1576--1585
  (2018)

\bibitem{lin2017focal}
Lin, T.Y., Goyal, P., Girshick, R., He, K., Dollar, P.: Focal loss for dense
  object detection. Proceedings of the IEEE International Conference on
  Computer Vision (ICCV)  (2017)

\bibitem{lin2014coco}
Lin, T.Y., Maire, M., Belongie, S., Hays, J., Perona, P., Ramanan, D., Dollár,
  P., Zitnick, C.L.: Microsoft coco: Common objects in context. Lecture Notes
  in Computer Science p. 740–755 (2014)

\bibitem{liu2016deepfashion}
Liu, Z., Luo, P., Qiu, S., Wang, X., Tang, X.: Deepfashion: Powering robust
  clothes recognition and retrieval with rich annotations. In: Proceedings of
  the IEEE Conference on Computer Vision and Pattern Recognition (CVPR) (2016)

\bibitem{liu2015deep}
Liu, Z., Luo, P., Wang, X., Tang, X.: Deep learning face attributes in the
  wild. In: Proceedings of the IEEE International Conference on Computer Vision
  (ICCV) (2015)

\bibitem{liu2020open}
Liu, Z., Miao, Z., Pan, X., Zhan, X., Lin, D., Yu, S.X., Gong, B.: Open
  compound domain adaptation. In: Proceedings of the IEEE Conference on
  Computer Vision and Pattern Recognition (CVPR) (2020)

\bibitem{liu2019largescale}
Liu, Z., Miao, Z., Zhan, X., Wang, J., Gong, B., Yu, S.X.: Large-scale
  long-tailed recognition in an open world. In: Proceedings of the IEEE
  Conference on Computer Vision and Pattern Recognition (CVPR) (2019)

\bibitem{mahajan2018exploring}
Mahajan, D., Girshick, R., Ramanathan, V., He, K., Paluri, M., Li, Y.,
  Bharambe, A., van~der Maaten, L.: Exploring the limits of weakly supervised
  pretraining. In: Proceedings of the European Conference on Computer Vision
  (ECCV). pp. 181--196 (2018)

\bibitem{mikolov2013distributed}
Mikolov, T., Sutskever, I., Chen, K., Corrado, G.S., Dean, J.: Distributed
  representations of words and phrases and their compositionality. In: Advances
  in Neural Information Processing Systems (NIPS). pp. 3111--3119 (2013)

\bibitem{ren2018learning}
Ren, M., Zeng, W., Yang, B., Urtasun, R.: Learning to reweight examples for
  robust deep learning. arXiv preprint arXiv:1803.09050  (2018)

\bibitem{shen2016relay}
Shen, L., Lin, Z., Huang, Q.: Relay backpropagation for effective learning of
  deep convolutional neural networks. In: Proceedings of the European
  Conference on Computer Vision (ECCV). pp. 467--482. Springer (2016)

\bibitem{tsoumakas2007overview}
Tsoumakas, G., Katakis, I.: Multi-label classification: An overview.
  International Journal of Data Warehousing and Mining (IJDWM)  \textbf{3}(3),
  1--13 (2007)

\bibitem{wang2016cnnrnn}
Wang, J., Yang, Y., Mao, J., Huang, Z., Huang, C., Xu, W.: Cnn-rnn: A unified
  framework for multi-label image classification. In: Proceedings of the IEEE
  Conference on Computer Vision and Pattern Recognition (CVPR). pp. 2285--2294
  (2016)

\bibitem{wang2017learning}
Wang, Y.X., Ramanan, D., Hebert, M.: Learning to model the tail. In: Advances
  in Neural Information Processing Systems (NIPS). pp. 7029--7039 (2017)

\bibitem{wang2017recurrently}
Wang, Z., Chen, T., Li, G., Xu, R., Lin, L.: Multi-label image recognition by
  recurrently discovering attentional regions. Proceedings of the IEEE
  International Conference on Computer Vision (ICCV)  (2017)

\bibitem{Xiong_2019_ICCV}
Xiong, Y., Huang, Q., Guo, L., Zhou, H., Zhou, B., Lin, D.: A graph-based
  framework to bridge movies and synopses. In: The IEEE International
  Conference on Computer Vision (ICCV) (2019)

\bibitem{zhang2007mlknn}
Zhang, M.L., Zhou, Z.H.: Ml-knn: A lazy learning approach to multi-label
  learning. Pattern Recognition  \textbf{40}(7),  2038--2048 (2007)

\bibitem{zhou2020BBN}
Zhou, B., Cui, Q., Wei, X.S., Chen, Z.M.: {BBN}: Bilateral-branch network with
  cumulative learning for long-tailed visual recognition. In: Proceedings of
  the IEEE Conference on Computer Vision and Pattern Recognition (CVPR).
  pp.~1--8 (2020)

\end{thebibliography}

\clearpage
\appendix
\section{Supplementary Materials}
\label{supplement}
\subsection{Dataset Construction}

\begin{figure}[hb] 
    \subfloat[\label{fig:pareto}]{\includegraphics[width=0.49\linewidth]{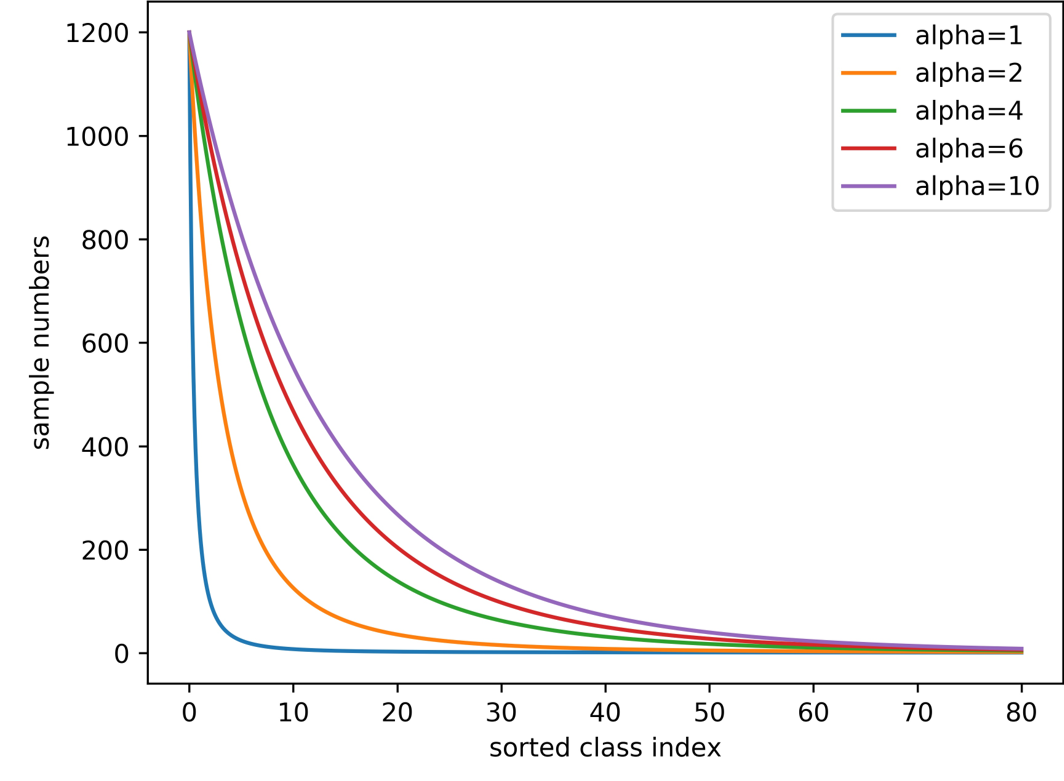}} \hfill
    \subfloat[\label{fig:construction}]{\includegraphics[width=0.49\linewidth]{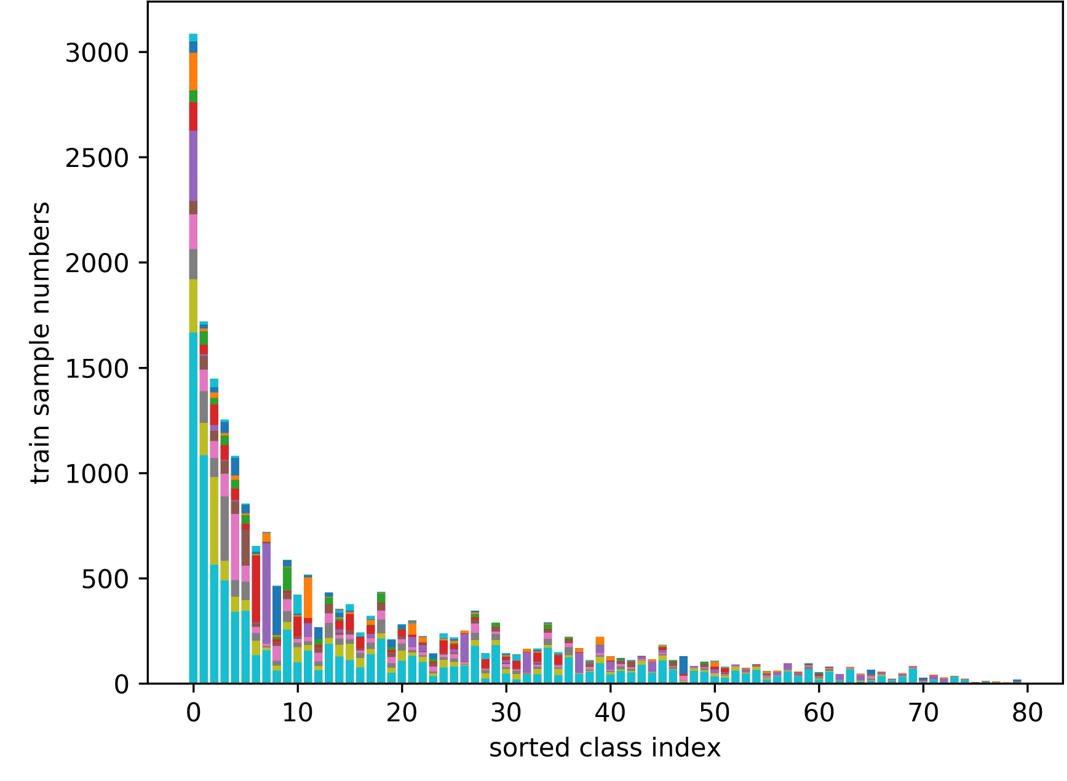}}
    \caption{
        (a). Pareto distribution with different $\alpha$.
        (b). The construction of COCO-MLT, where the colors visualize the increment of sample number per-class after each sampling iteration
    }
\end{figure}

\begin{figure}[t]
    \subfloat[\label{fig:voc_test_dist}]{\includegraphics[width=0.49\linewidth]{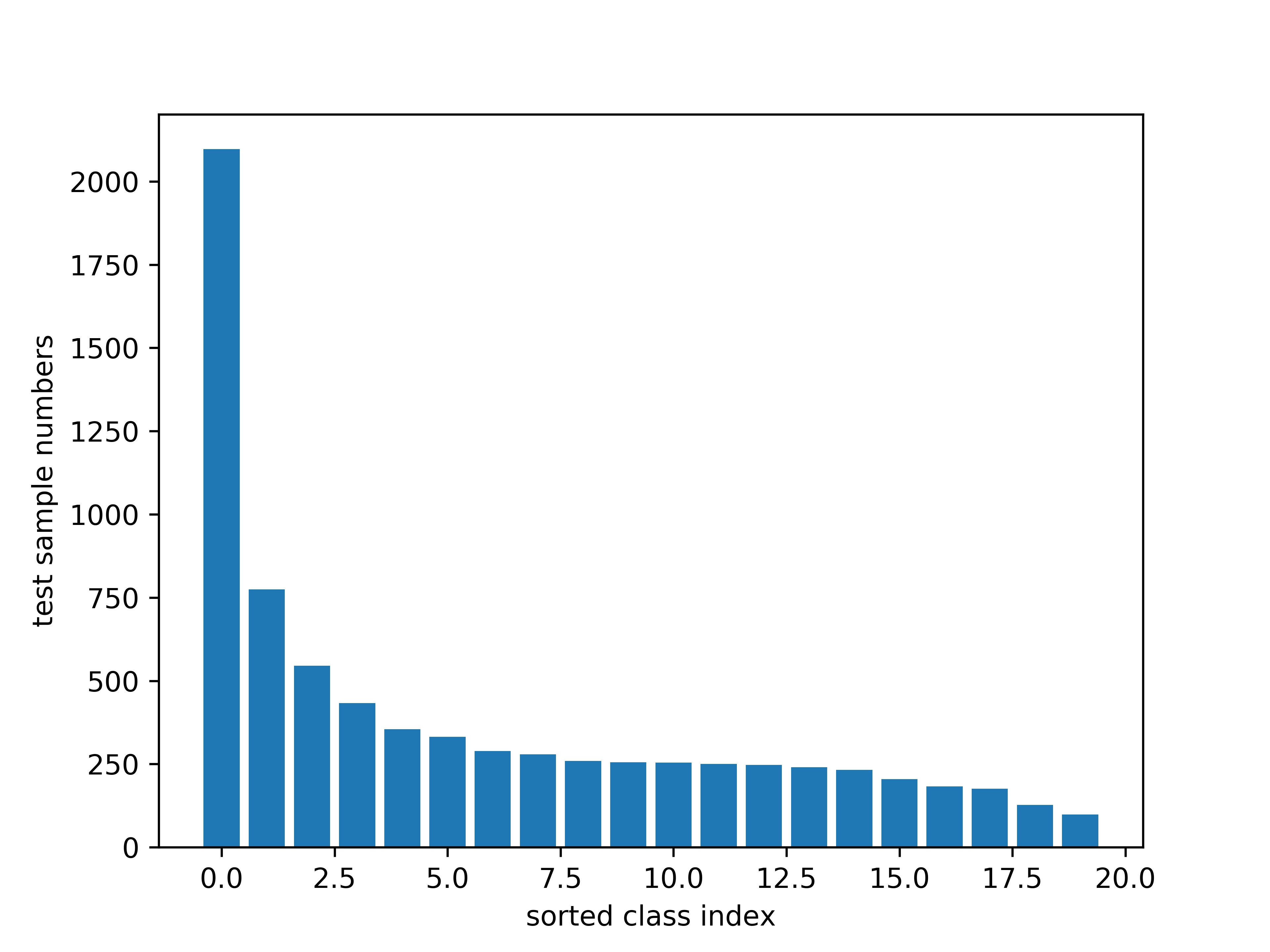}} \hfill
    \subfloat[\label{fig:coco_test_dist}]{\includegraphics[width=0.49\linewidth]{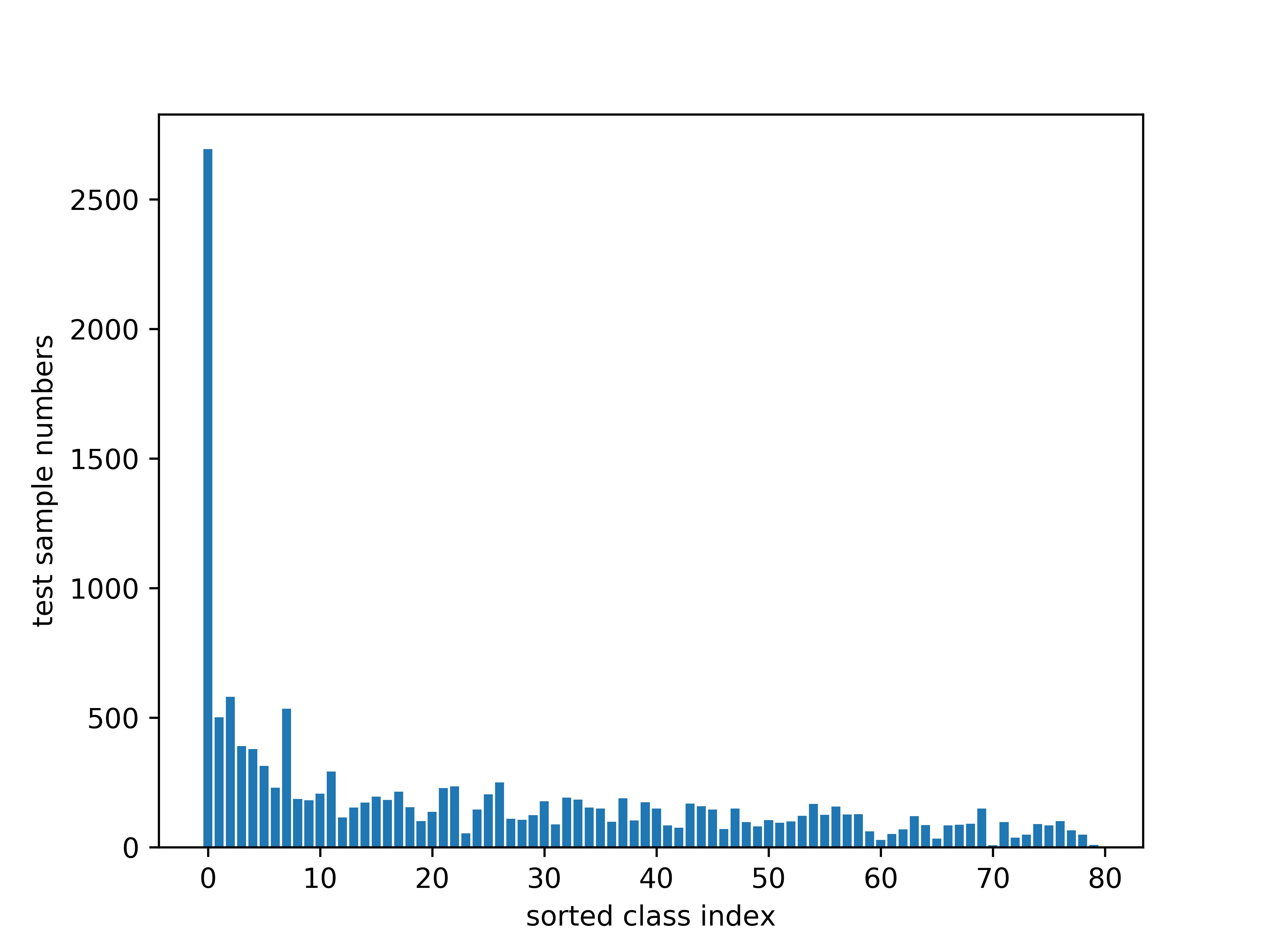}} 
    \caption{
        (a). The test set distribution of COCO2017, and we use the sorted sample number of the long-tailed training set as the x-axis index, which is relatively balanced except for one class(person).
        (b). The test set distribution of VOC2007
    }
    \label{fig:test_dist}
\end{figure}
We create our multi-label long-tailed datasets by extracting subsets from two multi-label image recognition benchmarks, VOC and COCO, respectively.
We adopt a \textit{pareto distribution} $pdf(x) = \alpha \frac{x_{min}^\alpha}{x^{\alpha + 1}}$ following Liu~\etal~\cite{liu2019largescale} with $\alpha$ controlling the shape of the distribution, as shown in Fig.~\ref{fig:pareto}. 
Concretely, we cut off the probability distribution function(pdf) when the cumulative distribution function(CDF) achieves 0.99, and then we rescale the pdf with a maximum of $N_{max}$, which is the maximum of sample numbers per class. Finally, we evenly split the x-axis into the number of classes as the original dataset, and we get a reference distribution.
We construct the datasets in a head-to-tail manner:
we first rank all the classes by $\hat{p}_i$ in Eq.1 mentioned in the main paper calculated with original data, and the subset is empty. For each class $i$ from head to tail, we compare the current sample number in the subset and the expected sample number by the reference distribution and then randomly \textit{add} or \textit{eliminate} certain instances accordingly. 
This way, we can constrain the tail classes to have a relatively small amount of data.
As seen in Fig.~\ref{fig:construction}, the construction is processed incrementally.
The distribution of the test set has a similar ranking order as the constructed train set, as shown in Fig.~\ref{fig:test_dist}. 
Except for one class, "person", the rest part of the test set is only slightly imbalanced.

\subsection{Implementation Details of Comparing Methods}
Some of the comparing methods are designed mainly for single-label datasets, and we make slight adjustments so that they work the best with our datasets:
For class-balanced(CB) loss~\cite{cui2019cb}, we set $\beta=0.99$ and $0.9$ for VOC-MLT and COCO-MLT, respectively, and we use an initial leanrning rate of 0.1, with an extra loss weight of 10 because we used an \textit{average} manner in the loss reduction while ~\cite{cui2019cb} used \textit{sum}; 
To calculate the effective numbers of a label set with multiple ground-truth, we adopt an average of $E_{n_i}$ calculated from each positive class, $\bar{E_n}=\frac{1}{\sum y_i} \sum_{i, y_i>0}{(1-\beta^{n_i})/(1-\beta)}$. 
We find it works better than $E_{\bar{n}}={(1-\beta^{\bar{n}})/(1-\beta)}$, where $\bar{n}=\frac{1}{\sum y_i}\sum_{i, y_i>0}n_i$.
For ML-GCN~\cite{chen2019mlgcn}, the dimension of the hidden layer in GCN is set as 256 and we use adjacent matrix generated from the long-tailed versions of datasets.
We also experiment with adjacent matrix generated from the original datasets so that it better matches the distribution of the test set, while the results show little difference.
For LDAM, we adopt a class-level margin, following~\cite{cao2019ldam} that we tune $C$ in $\frac{C}{n_i^{1/4}}$ and normalize the largest margin to be 0.5.

\subsection{The Effect of $\mu$ in Smoothing Function.}
In our paper, we report how the results are affected by $\beta$ of a smoothing function in Eq.5. And $\mu$ also controls the shape of the function in the actual range of variables.
As shown in Fig.~\ref{fig:smoothing_mu}, the influence is relatively small so we selected two insignificant peaks where $\mu = 0.2, 0.3$ for COCO-MLT and VOC-MLT, respectively, for the main experiments.
\subsection{The Effect of $\nu$ in Negative-tolerant Regularization}
\begin{figure}[t]
    \subfloat[\label{fig:smoothing_mu}]{\includegraphics[width=0.49\linewidth]{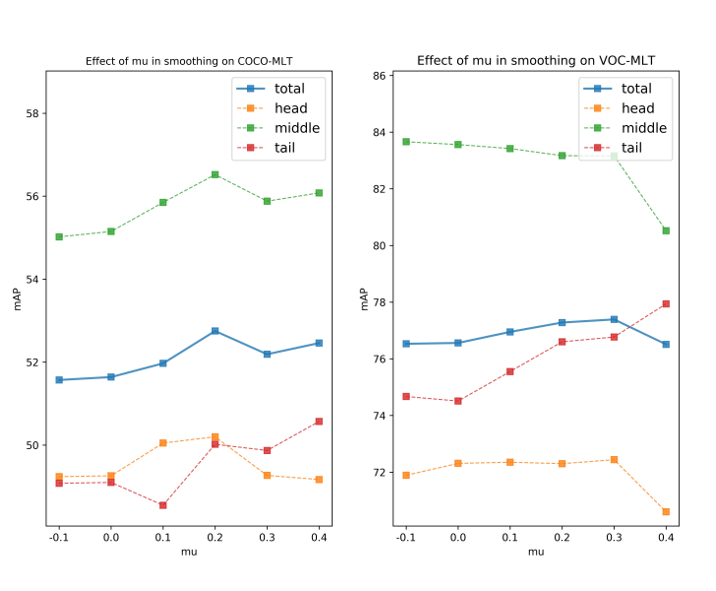}} \hfill
    \subfloat[\label{fig:nt_nu}]{\includegraphics[width=0.49\linewidth]{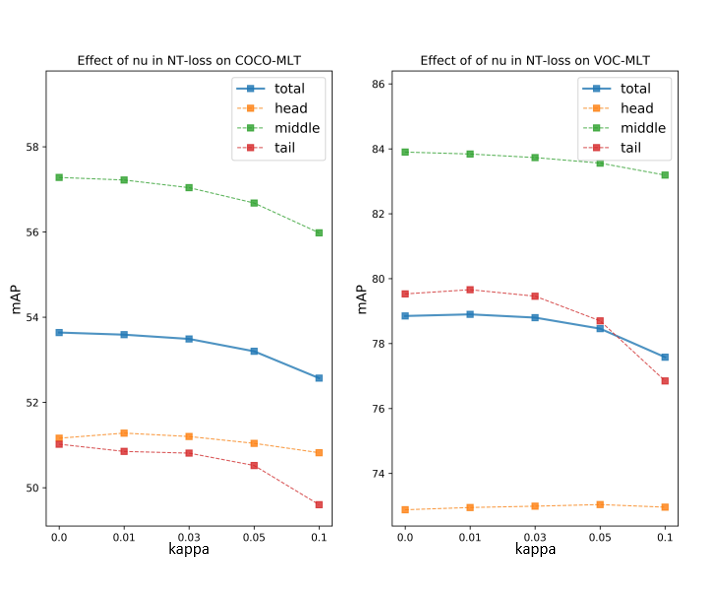}}
    \caption{
        (a). The effect of $\mu$ in the smoothing fuction.
        (b). The effect of $\kappa$ in netagive-tolerant regularization 
    }
\end{figure}

To under stand how $\nu$ the netagive-tolerant regularization in Eq.12 in affect the results independently, we fix $\lambda=2,5$ for COCO-MLT and VOC-MLT, respectively, and change $\nu$ by changing $\kappa$, as shown in Fig.~\ref{fig:nt_nu}.
Setting $\kappa=0$ has a relatively good result, which means that the thresholds in the regularization can be simply fixed as zero, and changing $\kappa$ in a small range has little effect.

\end{document}